\documentclass[11pt,a4paper]{article}
\usepackage{authblk}
\usepackage[hyperref]{acl2021}
\usepackage{times}
\usepackage{latexsym}
\usepackage{framed}
\usepackage{amsmath} 
\usepackage{graphicx}
\usepackage{dsfont}
\usepackage{amsfonts}
\usepackage{subfigure}

\usepackage{microtype}

\aclfinalcopy 
\def\aclpaperid{3176} 

\title{Can Transformer Models Measure Coherence In Text?\\ Re-Thinking the Shuffle Test}

\author[ ]{Philippe Laban}
\author[ ]{Luke Dai}
\author[ ]{Lucas Bandarkar}
\author[ ]{Marti A. Hearst}
\affil[ ]{UC Berkeley}
\affil[ ]{\texttt{\{phillab, luke.dai, lucasbandarkar, hearst\}@berkeley.edu}}
\date{}

\begin{document}
\maketitle


\begin{abstract}
The \textit{Shuffle Test} is the most common task to evaluate whether NLP models can measure coherence in text.
Most recent work uses direct supervision on the task; we show that by simply finetuning a RoBERTa model, we can achieve a near perfect accuracy of 97.8\%, a state-of-the-art.
We argue that this outstanding performance is unlikely to lead to a good model of text coherence, and suggest that the Shuffle Test should be approached in a Zero-Shot setting: models should be evaluated without being trained on the task itself. We evaluate common models in this setting, such as Generative and Bi-directional Transformers, and find that larger architectures achieve high-performance out-of-the-box. Finally, we suggest the k-Block Shuffle Test, a modification of the original by increasing the size of blocks shuffled. Even though human reader performance remains high (around 95\% accuracy), model performance drops from 94\% to 78\% as block size increases, creating a conceptually simple challenge to benchmark NLP models.
\end{abstract}

\section{Introduction}

In recent years, text generation applications, fueled by Transformer models pre-trained on large datasets, have achieved dramatic results on a wide range of NLP tasks.  These include  GPT2 applied to story completion of fan fiction \cite{radford2019language},  the PEGASUS model \cite{zhang2020pegasus} improving state-of-the-art on ten summarization datasets in widely varying domains, and more recently  GPT3 \cite{Brown2020LanguageMA} doing well on a diversity of tasks in a zero-shot setting. However, it is not clear how \textit{coherent} the text generated by these models is.

\begin{figure}
    \centering
    \includegraphics[width=0.5\textwidth]{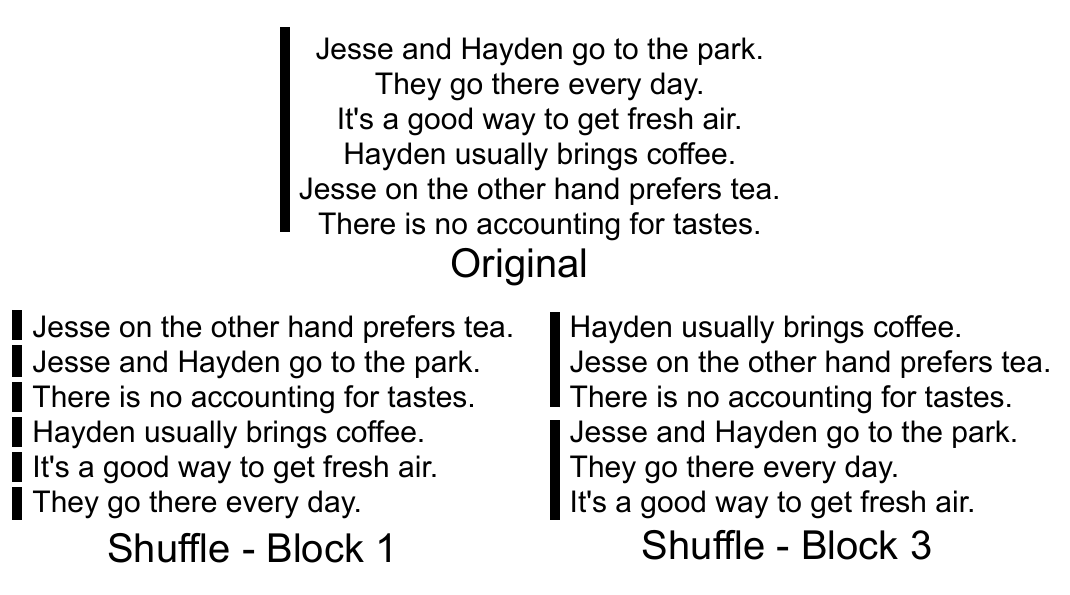}
    \caption{\textbf{Can modern NLP models recognize shuffled, incoherent text without supervision?} Yes (mostly) when all sentences are shuffled (left), but less so when shuffling $k$ blocks at a time (right).}
    \label{fig:block_shuffle_example}
\end{figure}

The computational linguistics literature holds many competing definitions of \textit{coherence} in text;  \citet{wang2014short} provide a useful brief summary of key competing theories.
This work attempts to identify the \textit{absence} of coherence, noting that a text might be composed of valid sentences when viewed independently, but when read sequentially, semantic relations are not well-supported.


The NLP community has proposed models to \textit{measure} coherence, as well as repeatable tasks to evaluate these models.
In this paper, we outline these common tasks, and describe what we believe is a limitation in the framework of the most common task: the Shuffle Test.
The Shuffle Test is a conceptually simple and reproducible task, in which a model must differentiate between an original text and a sentence-order shuffled version.
Because of its simplicity, we make the argument that the Shuffle Test should be viewed as a \textit{probe}: a task on which models should be evaluated without directed supervision.
Prior work \cite{paulus2018deep} has shown that directly optimizing evaluation metrics such as ROUGE or BLEU leads to inadequate models, exploiting weaknesses in the evaluation metrics.

We show that this phenomenon occurs with the current application of the Shuffle Test in related work. To demonstrate the pitfalls, we finetune a RoBERTa-large model -- an architecture several orders of magnitude larger than previously used models -- on the Shuffle Test and show the results outperform previous models, with an accuracy of 97.8\%. We argue this model has most likely learned features specific to recognizing shuffled-ness, which is probably a conflated signal for the underlying goal of  learning a strong coherence model.

We first outline prior work on tasks and models to measure textual coherence, then describe the framework for the Zero-Shot Shuffle Test, showing how to adapt common models to the setting, and finally propose a variation to the Shuffle Test that significantly increases the challenge for models, while not affecting human performance at the task.\footnote{The code and model checkpoints are available at: \url{https://github.com/tingofurro/shuffle_test}.}


\section{Tasks and Models for Coherence}

\begin{figure}
    \centering
    \includegraphics[width=0.47\textwidth]{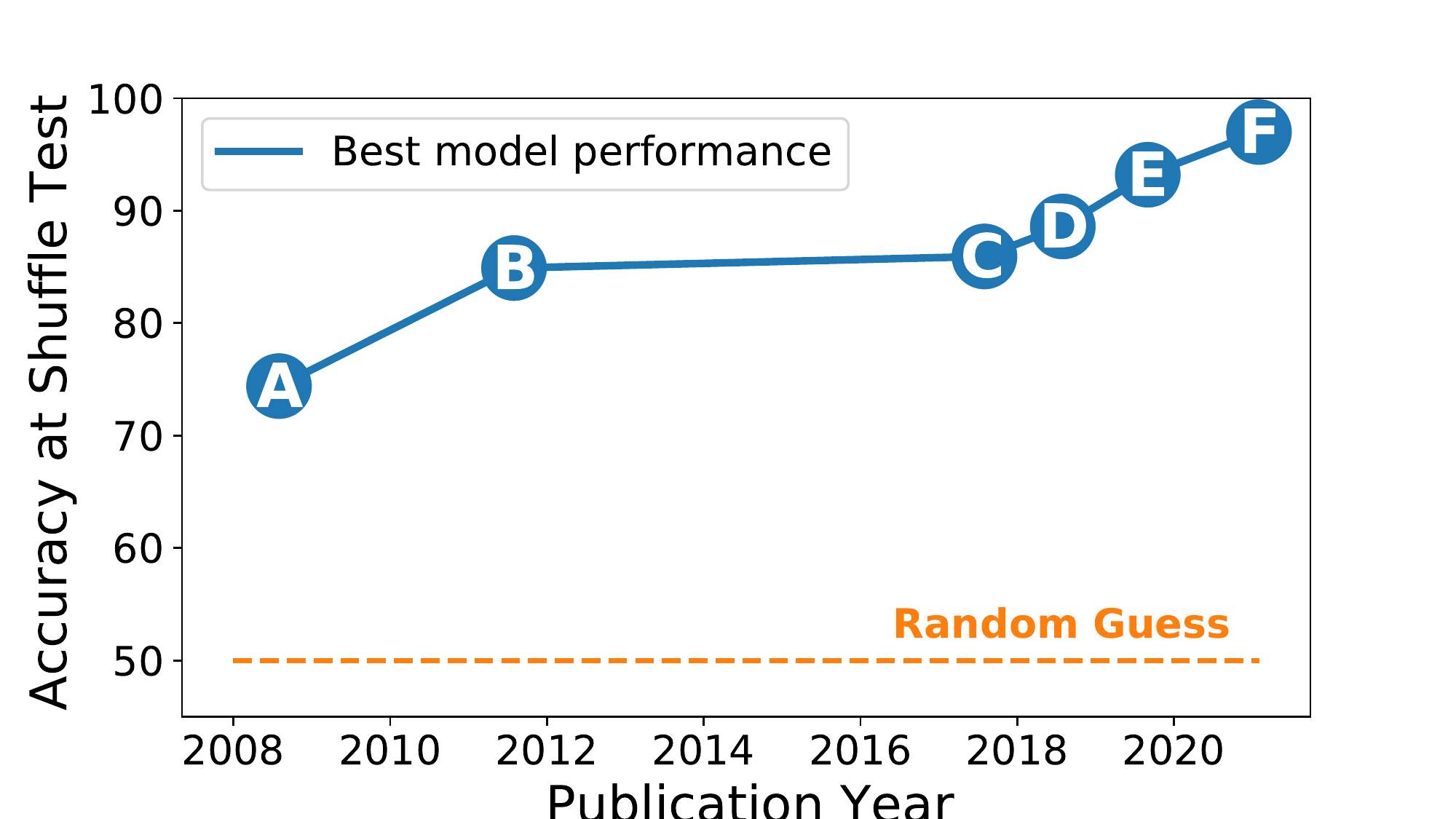}
    \caption{Timeline of incremental accuracy improvements on the \textit{Shuffle Test} on the WSJ corpus. Letters are for models described in Section~\ref{sec:supervised_models}.}
    \label{fig:supervised_shuffle_timeline}
\end{figure}

\subsection{Tasks for Coherence Evaluation}

\textbf{The Shuffle Test}, introduced by \citet{Barzilay2008ModelingLC}, is the most common task for coherence model evaluation. The task is a binary classification, in which a model must discriminate between a document and a \textit{shuffled document}, obtained by randomly shuffling the order of sentences in the document. The most common dataset for evaluation is a set of articles from the Wall Street Journal \cite{elsner-charniak-2011-extending}.

In \textbf{the Insertion Test}, a single sentence from a document is removed, and the model must predict the sentence position. Typically, models assign a score to each possible position, and predict the one with highest score. One limitation of the Insertion Test is that model accuracies are low, often in the 10-20\% range \cite{elsner-charniak-2011-extending}. To our knowledge, there is no evaluation of human performance on this test; with the possibility that the task can have more than one plausible solution. Computational cost is another limitation, often growing linearly with the number of sentences.

In \textbf{the Sentence Ordering Task}, a model is given an randomly ordered sentence set, and must produce the correct ordering of sentences. The task is often restricted to generative models, as it is prohibitively expensive to score all combinations to extract a best-scoring order \cite{logeswaran2018sentence}.


\subsection{Models for the Shuffle Test}
\label{sec:supervised_models}
Figure~\ref{fig:supervised_shuffle_timeline} is a timeline of models that have led to progress on the Shuffle Test since its introduction.

The Entity Grid (model A in Fig.~\ref{fig:supervised_shuffle_timeline}) was introduced by \citet{Barzilay2008ModelingLC}. A text is transformed into an entity grid, a matrix (\#sentences x \#entities) indicating presence of an entity in a sentence. The entity grid is featurized and used to train a predictor on coherence tasks.

\citet{elsner-charniak-2011-extending} (model B) extended the entity grid by adding linguistic features such as named-entity type. \citet{nguyen2017neural} (model C) introduce the first neural approach, using a convolutional neural network (CNN) to operate over the entity grid, and \citet{joty2018coherence} (model D) added word embeddings to entity-grid features. Most recently, \citet{moon2019unified} (model E) replaced traditional word vectors with ELMO \cite{peters2018DeepCW} contextual word vectors.

Crucially, all these models are directly trained on the Shuffle Test, and with each iteration of improvement, model capacity (i.e., the number of trainable parameters) has increased. We finetune a RoBERTa-large \cite{liu2019roberta} (model F), a still larger model, on the Shuffle Test, and achieve a 97.8\% accuracy on the WSJ test set, a new state-of-the-art. 

\textbf{Training details}. We finetune the RoBERTa-large on the training portion of the WSJ dataset, and setup the task as a sequence classification. We trained using the ADAM optimizer, using a learning rate of $1e^{-5}$ and a batch-size of 16. The model was trained on a single GPU, an Nvidia V-100, and training converged within 10 minutes. Model checkpoint was selected based on a validation set accuracy, and tested once on the standard WSJ test set.

This stellar performance leads us to believe that there are two conflated factors that cause good performance on the Shuffle Test: a model that can truly recognize the lack of coherence in shuffled text, and a model specialized at the Shuffle Test, recognizing shuffle-specific features in text, without assessing textual coherence. This resonates with findings from \citet{mohiuddin2020coheval}, who show that increased model performance at the Shuffle Test in supervised models does not necessarily lead to improvements in downstream tasks, such as ranking of generated summaries.  Ideally, the Shuffle Test would be used to assess coherence models that work independently of the test itself. 


We propose a simple solution: coherence models should be evaluated on the Shuffle Test in a Zero-Shot setting; without being supervised on the test, preventing the learning of shuffle-specific features, and more directly evaluating coherence aptitudes.

\section{Zero-Shot Shuffle Test}
\label{sec:zero_shot}

We now define specifically what factors need to be respected to satisfy the zero-shot setting.

\begin{quote}
    In the Zero-Shot Shuffle Test, the evaluated model must not be pre-trained, finetuned or modified using shuffled text.
\end{quote}

More specifically, this restricts the use of the Shuffle Test as supervision (as a binary classification task), as well as other tasks that involve shuffling text, such as the sentence ordering task.

Next, we adapt common architectures to the Zero-Shot Shuffle Test and assess performance in diverse textual domains.

\subsection{Zero-Shot Coherence Models}

We adapt the two common classes of Transformer models to the Zero-Shot Shuffle Test: Generative and Bi-directional Transformers.

\textbf{Generative Transformers} are compatible with  language modeling, in which a model assigns a likelihood to a sequence of words ($S = [W_1, .... W_n]$). Transformers estimate the likelihood of a sequence by factoring on sequence order:
\begin{equation}
    P(S) = \prod_{i=1}^N P(W_i | W_1 ... W_{i-1})
\end{equation}
Taking the log of the likelihood ($log(P(S))$) is  often preferred as it allows for numerical stability.

To perform a Zero-Shot Shuffle Test, we compute log-likelihoods of the original and shuffled documents and predict the lower-scoring one as shuffled.

We  experiment with GPT2 models of varying sizes (\textbf{GPT2-base}, \textbf{GPT2-medium}, \textbf{GPT2-large}), and finetune an In-Domain GPT2-medium using a language modeling loss in each domain to evaluate whether in-domain specialization improves performance (\textbf{GPT2-med-ID}).

When texts exceed sequence-length limits of models (e.g., 512 words), we implement a sliding window mechanism. The sequence is split into successive windows with 50\% overlap. Window log-likelihoods are averaged into a document log-likelihood.



\textbf{Bi-Directional Transformers}, exemplified by BERT \cite{devlin2019bert}, are the second class of models we adapt to the test. Unlike Generative Transformers, bi-directional Transformers do not impose strict sequence order, rendering sequence likelihood estimation less straightforward.

\citet{salazar2020masked} propose a solution, with Masked Language Model Scoring ($\mathrm{MLMS}$), in which a likelihood is estimated by masking each word in the sequence, predicting its identity, and averaging all word-likelihoods into a score:
\begin{equation}
    \mathrm{MLMS}(S) = \frac{1}{N} \sum_{i=1}^N \log P_{MLM} (W_i | W_{\backslash i})
\end{equation}
where $W_{\backslash i}=S-\{W_i\}$. Unlike generative models, each word's likelihood is conditioned on all others, an advantage of Bi-directional models. For the Zero-Shot Shuffle Test, the document with lower $\mathrm{MLMS}$ is predicted as shuffled.

One key disadvantage of $\mathrm{MLMS}$ is its computational cost: scoring requires a forward-pass for each word in the sequence; by contrast, generative models usually require a single forward pass. This limits our ability to test large models, and therefore test only base models: \textbf{BERT-base} and \textbf{RoBERTa-base}.


\subsection{Datasets}

To examine whether there are significant differences in performance across domains, we evaluate with the Shuffle Test using three distinct domains. We performed a manual check to determine that the datasets we selected do not overlap with the dataset used to pre-train BERT, RoBERTa and GPT2. The three domains are:

\textbf{News domain}. We use the standard Wall Street Journal (WSJ) test-set introduced by \citet{elsner-charniak-2011-extending} in the supervised Shuffle Test. The dataset contains 1006 documents.

\textbf{Legal domain}. We use the full document released in the BillSum dataset \cite{kornilova2019billsum} which consists of US Congressional and California state bills. We use the first 1,000 documents in the standard test set.

\textbf{Blog domain}. We use posts of the Reddit TIFU dataset \cite{kim2019abstractive}, consisting of stories written by members of the Reddit community. We use the first 1,000 documents of the corpus.

We choose these datasets as they are publicly available, can easily be accessed through the HuggingFace datasets package \cite{2020HuggingFace-datasets} and represent a diversity of textual domains.

We note that document length affects the amount of displacement that occurs from shuffling, with more displacement in longer texts. To take this effect into account, we truncate documents at 20 sentences before administering the Shuffle Test.

\subsection{Results}

\begin{table}[]
    \centering
    \resizebox{0.45\textwidth}{!}{%
    \begin{tabular}{lrrrrr}
                       & \multicolumn{4}{c}{\textbf{Domain (\%)}} \\ \hline
    \textbf{Model}     & \textbf{WSJ} & \textbf{Legal} & \textbf{Reddit} & \textbf{Overall} \\ \hline
    GPT2-base    & 47.2 & 92.0 & 74.8 & 71.3 \\
    GPT2-medium  & 91.2 & 98.6 & 88.9 & 92.9 \\
    GPT2-large   & 73.2 & \textbf{99.3} & 90.6 & 87.7 \\
    BERT-base    & 73.2 & 96.1 & 86.1 & 85.1 \\
    RoBERTa-base & 82.3 & 94.8 & \textbf{96.7} & 91.3 \\
    GPT2-med-id  & \textbf{93.1} &  98.8 & 90.0 & \textbf{94.0} \\ \hline
    \end{tabular}
    }
    \caption{Accuracy of Zero-Shot Shuffle Tests of models on three domains: Wall Street Journal (WSJ), Billsum documents (Legal), and Reddit. We report an overall, averaged performance across domains.}
    \label{tab:zero_shot_results}
\end{table}

Overall, all models significantly outperform random chance, with the GPT2-medium achieving 91.2\% on the WSJ test-set out of the box where the previous supervised state-of-the-art was 93\%.
Bi-directional models achieve better results than generative models at Transformer-base size (e.g., GPT2-base vs. RoBERTa-base). 

Increasing model size leads to large performance improvement for GPT2, confirming that according to the Shuffle Test, larger Transformer models improve at modeling coherence. In-domain finetuning leads to an improvement on all domains (GPT2-med-id outperforming GPT2-medium), confirming the strength of in-domain finetuning \cite{howard2018universal}.

Finally, models achieved stronger performance on the Legal domain, with models all scoring 92.0 or above. Overall, three of the six models we test achieve compound performance over 90\%. 

There is a potential question about the zero-shot nature of the BERT training method. The original BERT model is trained with two objectives, one of which is Next Sentence Prediction (NSP). In NSP, the model is exposed to two blocks of text, and must predict whether they are adjacent in a document or not. It can be argued that NSP is an indirect supervision signal for the Shuffle Test. However, we find that the BERT model performs worse than RoBERTa, a similar model in architecture trained without the NSP objective. This difference in performance suggests that the NSP objective is not the cause of the superior performance of these models, thus preserving the claim that they act in a zero-shot manner for the purposes of the Shuffle Test.

We next propose a modification to the Shuffle Test that challenges models significantly more.

\section{The $k$-Block Shuffle Test}

\begin{table}[]
    \centering
    \resizebox{0.47\textwidth}{!}{
    \begin{tabular}{lccccc}
                      & \multicolumn{5}{c}{\textbf{Block Size}}                        \\ \hline
    \textbf{Model}    & \textbf{1} & \textbf{2} & \textbf{3} & \textbf{4} & \textbf{5} \\ \hline
    Human Perf. - WSJ & 97.5       & 94.5       & 93.0       & 96.0       & 94.0       \\ \hline
    GPT2-med - WSJ    & 95.3       & 91.4       & 89.5       & 87.4       & 85.3       \\
    GPT2-med - Legal  & 98.7       & 98.0       & 96.9       & 95.9       & 94.5       \\
    GPT2-med - Reddit & 89.5       & 76.9       & 66.1       & 59.1       & 53.8       \\
    GPT2-med - Avg.   & 94.5       & 88.8       & 84.2       & 80.8       & 77.9       \\ \hline
    \end{tabular}
    }

    \caption{\textbf{Results of Zero-Shot KBST varying the block size from one to five.} The GPT2-medium model was tested on all three domains, and human performance was measured on WSJ.}
    \label{tab:block_results}
\end{table}

Results in Section~\ref{sec:zero_shot} can be interpreted to mean that with large enough Transformer models, the Shuffle Test with no supervision is essentially a solved task.
We find that a simple modification of the Shuffle Test can significantly reduce model performance, without affecting human annotator performance.

The modification we propose, k-Block Shuffle Test (KBST), is illustrated in Figure~\ref{fig:block_shuffle_example}. In the standard Shuffle Test, text is divided into sentences and shuffled, with a unit of one sentence. In the k-Block Shuffle Test, sentences are grouped in contiguous blocks of k sentences (resembling paragraphs), and the blocks are shuffled, maintaining sentence order in each block.
Within a block, sentences remain locally coherent, and as block size increases, the fraction of correct sentence transitions increases, while potentially incoherent transitions decrease.

To establish the feasibility of the KBST with differing block-sizes, we performed a human evaluation completed by authors of the paper as well as a third annotator recruited on the Upwork\footnote{https://www.upwork.com} platform.  This annotator is a native English speaker with experience in proofreading, and was remunerated at \$15/hour USD.

The human evaluation consisted of performing KBST on 500 documents randomly sampled from the WSJ dataset, with 100 tests for each block-size from one to five. Each Shuffle test was performed by at least two annotators. We find that there is high inter-annotator agreement (Cohen's Kappa $\kappa = 0.86$), which does not significantly vary with block-size (ranging from 0.76-0.94).

KBST results for human and computational models are shown in Table~\ref{tab:block_results}.
Human performance is very high on WSJ, averaging above 95\%, and is not significantly affected by block-size.

Timings logged during human annotation show that Shuffle Tests took on average 40\% more time for larger (3-5) than smaller blocks (1-2), showing the task requires more attention from annotators as block size increases.

In all three domains, increased block size leads to a decrease in model performance. The magnitude of decrease in performance from block-size 1 to 5 is sensitive to the domain, with a drop of 4\% in the legal domain, and 36\% for Reddit, on which the 5-block performance of 53.8\% narrowly outperforms random performance.

Aggregate model performance drops from 94.5\% for block-size 1 to 77.9\% for block-size 5, leaving significant room in larger block-size to measure future model improvements.

Although increasing the block size leads to a more challenging task for current models, we argue models should not be evaluated on a single block size, but on  several block sizes, with each block size giving a perspective on the model's performance at a specific point between local and global coherence \cite{vandijk1985semantic}.


\section{Limitations and Future Work}

\textbf{Shuffling vs. Coherence.} In this work, we propose an improved setting for the Shuffle Test, the most popular probe to measure textual coherence. However, many linguistic phenomena necessary for coherence of text cannot be measured by shuffling sentence order. In the long-run, the community should build more elaborate coherence measures, to build a more complete picture of model capabilities and limitations.

\textbf{Coherence in Long Text.} We limited our analysis to texts with up to 512 words, a common constraint in pre-trained Transformers. Recent progress in model architectures open the possibility to process longer text, with models such as the Reformer \cite{kitaev2019reformer}, Longformer \cite{Beltagy2020Longformer} and Big Bird \cite{zaheer2020bigbird} processing sequences of several thousand words. With longer sequences, one can further increase the block-size of the k-Blocked Shuffle Test (i.e., k=20) to gain understanding of model's ability to discern global coherence \cite{vandijk1985semantic}, or main topics and subtopics \cite{hearst1997text}.

\textbf{Specialized Coherence Models.} In this work, we limit our analysis to popular models out-of-the-box, establishing baseline performances for the Zero-Shot KBST. Future work should establish whether performance can be further improved, for instance using word-level likelihood signals and surprisal profiles.

\section{Conclusion}

In this work, we discuss a potential limitation in the framing of the Shuffle Test, the most commonly used task to evaluate models for textual coherence. We show that a RoBERTa model can be finetuned to achieve near-perfect performance without necessarily measuring coherence, and propose a new framework: the Zero-Shot Shuffle Test, in which direct supervision is disallowed. Modern NLP architectures can achieve high performance out-of-the-box in this Zero-Shot setting on a variety of textual domains. We find however that models struggle when we introduce a simple modification, k-Blocking, to the shuffling strategy, with accuracy dropping from around 94\% to around 78\%. The k-Block Shuffle Test in a Zero-Shot setting is a straightforward, reproducible task that can be used to benchmark future NLP architectures to measure coherence capabilities.

\section*{Acknowledgments}

We would like to thank Katie Stasaski, Dongyeop Kang, and the ACL reviewers for their helpful comments. This work was supported by a Microsoft BAIR Commons grant as well as a Microsoft Azure Sponsorship.

\section*{Ethical Considerations}

We present a method to evaluate models on their ability to measure textual coherence. We have exclusively run experiments in the English language, and even though we expect the method to be adaptable to other languages, we have not verified this assumption experimentally and limit our claims to the English language.

For the human evaluation, we paid the annotator above the minimum wage, and do not release any personal identifiable information. We did not collect payment information and relied on a third party (Upwork.com) for remuneration.

\bibliographystyle{acl_natbib}
\bibliography{anthology,acl2021}

\end{document}